\documentclass{article}

\usepackage{threeparttable}
\usepackage{arxiv}
\usepackage[utf8]{inputenc} 
\usepackage[T1]{fontenc}    
\usepackage{hyperref}       
\usepackage{url}            
\usepackage{booktabs}       
\usepackage{amsfonts}       
\usepackage{nicefrac}       
\usepackage{microtype}      
\usepackage{lipsum}
\usepackage{graphicx}
\graphicspath{ {./images/} }
\usepackage{cite}

\title{A SOLUTION TO PRODUCT DETECTION IN DENSELY PACKED SCENES}

\author{
 Tianze Rong, Yanjia Zhu, Hongxiang Cai, Yichao Xiong \\
  \texttt{dominirong@gmail.com, 13218012112@163.com, hxlll@126.com ,xyc\_sjtu@163.com}
}

\begin{document}
\maketitle

\section{Introduction}
Densely packed scene detection is a extension of the object detection, but with a larger number, denser arrangement. We adopted a modified random crop strategy and a optimized Cascade R-CNN to solve the problems. Finally, we achieved a mmAP as 58.7\% on SKU-110k\cite{sku}. 
\section{Method}
\subsection{Data Description and Analysis}
\paragraph{Statistics of Image} After eliminating the invalid data, the size of the data set is shown below:
\begin{table}[htb]
    \centering
    \begin{tabular}{ccc}
    \hline
        &Image Number & Annotation Number \\
    \hline
         Train Set& 8219& 1208482\\
         Validation Set & 588 & 90968\\
         Test Set1& 2936 & 431546\\
    \hline
        Total & 11743 & 1730996\\
    \hline
    \\
    \end{tabular}
    \caption{Quantitative Statistics of SKU-110k}
    \label{tab:my_label}
\end{table}
\par Dense object scene is mostly with a huge number of objects in a single image. The table below states the statistics of number of annotations in a single image:
\begin{table}[htb]
    \centering
    \begin{tabular}{cccccc}
    \hline
        &Mean & Max & Min & 99.5\% percentile & 0.5\% percentile\\
    \hline
         Train Set& 147& 576&1&356&61\\
         Validation Set & 154 & 759&40&582&59\\

    \hline
    \\
    \end{tabular}
    \caption{Statistics of Number of Annotations in Single Image}
    \label{tab:my_label}
\end{table}
\paragraph{Statistics of Annotations} Assume the input size is (1333,800), and the scale of a single bounding box represents as $\sqrt{wh}$, where $w,h$ stand for the width and height of bounding box. Here, the figure 1, is the histogram and cumulative ratio curve of the scales of the data set.
\begin{figure}[htb]
    \centering
    \includegraphics[scale=0.6]{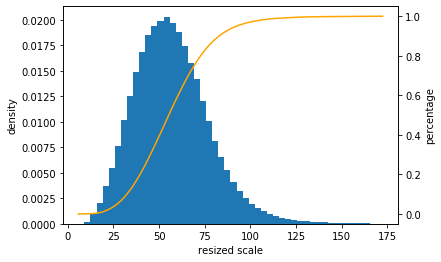}
    \caption{the Histogram and Cumulative Ratio Curve of the Scales}
    \label{fig:my_label}
\end{figure}

\par As we can see, a third of bounding box is small object by the definition of MS COCO data set\cite{coco} after rescaling, the scale of the object is smaller than 32 pixels.
\subsection{Solution to Densely Packed Scene}
\subsubsection{Promotion via Data}
\paragraph{Rescale}As the analysis before, it is apparent that small object problem is a vital factor should be solved to keep a better performance of model. The naivest solution to small objects is to enlarge the input size of images, which is able to make the small objects into larger ones. Regarding that SKU-110K data set possesses a high resolution level, a large input size would not take the side-effect significantly.
\paragraph{Random Crop} Theoretically, we should input the images as large as possible. However, practically, the GPU memory capacity usually limits the input size in most of the cases. We adopt random crop to combines these two urges. But something need to be paid more attention here:
\begin{itemize}
    \item When bounding boxes right on the cropping border, it would be clipped and remain the reserved region.
    \item Random sampling is probable to make some of the objects never be sampled. 
\end{itemize}
\subparagraph{Random Seven Crop} We designed a strategy to relieve these two disadvantage. Clipping the bounding box may cause some fake box whose entity in box has been clipped out but background still remains. These fake box can lead to confusion to model in training. Hence we only remain a clipped box whose IoU to origin box higher than a threshold. Regular random crop sample the position of crop region from a uniform distribution. Random Seven Crop is designed to sample the region from only seven certain position: Four corners of image, center point and two end points of short axis.
\begin{figure}[htb]
    \centering
    \includegraphics[scale=0.4]{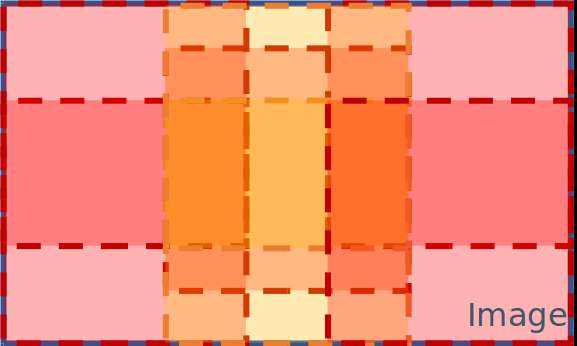}
    \caption{The seven sample areas of Random Seven Crop}
    \label{fig:my_label}
\end{figure}
\subsubsection{Adjustment of Detector}
\paragraph{Sampler Hyper-parameters}Besides small object problem, another problem is about counts of the bounding boxes in a single image. As the statistics, a single images has a average as about 150 boxes, a number far more than MS COCO. But most of the default hyper-parameters are based on MS COCO data set which has a lower density of objects than SKU-110K. Hence we adjust the max positive sample number of both RPN and R-CNN sampler to release the limits.
\paragraph{Cascade R-CNN} The model can achieve a relatively high AR@50 after sorts of optimization, while mAP(0.5:0.95) declined rather rapidly from AP@50. Cascade R-CNN\cite{cascade} can refine those bounding box whose location is not that accurate by cascading bounding box heads. In this case, more accurate localization not only makes tight bound, but suppresses duplicated boxes so that reduces false positive boxes.
\subsection{Other Modification}
\paragraph{Inference Hyper-parameters} Inference hyper-parameters is occasionally neglected by developers during improvement, in this case a.k.a NMS hyper-parameters optimizing. We adopt grid search eliminate by orthogonal design to search the optimal combination of the hyper-parameter.
\paragraph{Backbone and Neck} Empirically, larger networks with more parameters bring better performance. To make an overview on the architecture of two-staged detector with neck. Backbone is the one of the easiest module to be adjust. So we replace the ResNet to ResNeXt\cite{resnext}. Neck is also available now to be replaced as easy as backbone. BFP\cite{bfp} is a neck structure with a balanced integration between octaves, meanwhile it can broadly improve the performance of the model. 

\section{Experiment}
All experiment are conducted on MMDetection Platform with single GPU and run evaluation by library pycocotools on validation set. Since that, all mmAPs presented in this work are COCO-style and the IoU threshold of mmAP is [0.5:0.95:0.05]. Because of the differences between SKU-110K and MS COCO, we set the maxDet parameter of pycocotools on 400 instead of the default setting as 100.
\subsection{Baseline}
We built a baseline via Faster R-CNN\cite{faster} with FPN\cite{fpn} and ResNet-50\cite{resnet} as backbone, also all configures are inherited from the Faster R-CNN with FPN benchmark on MS COCO from MMDetection\cite{mmdet}. The result mmAP is 50.6\%, which is shown in the first row of the result table.
\subsection{Performance Improving of Inference}
We designed a orthogonal experiment to optimize NMS parameters using $L_9(3^4)$. As for the result of orthogonal table, we analyzed with ANOR. And the optimal parameters combination is following:
\begin{itemize}
    \item Input and Output boxes number = 3000
    \item Minimal confidence threshold = 0.05
    \item IoU threshold of positive sample = 0.7
    \item Max output boxes number = 400
\end{itemize}
\subsection{Adjustment on Sampler of Detector}
According to the statistics, we already knew that number of ground truth in single image impacts the model vitally. We aim on the samplers among the all modules in detector by ruling the others out one by one, since sampler is one of the strongest modules coupled with the quantitative characteristics of a data set. The default RPN sampler number is 256 on COCO and there is a fraction of 0.5 on positive sample, which is not enough for SKU-110K apparently. And we counted positive sample number in all image. The result is illustrated in figure 3. So we enlarge the sample number to 512. Furthermore, we set the R-CNN sampler number to 3072 in accordance with RPN, since we got the called for conclusion on R-CNN sampler.
\begin{figure}[htb]
    \centering
    \includegraphics[scale=0.6]{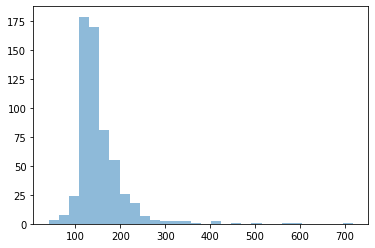}
    \caption{The histogram of positive sample number in a single image(x axis is positive sample number, y axis is count of this bin)}
    \label{fig:my_label}
\end{figure}
\subsection{Random Crop and its Modification}
The regular random crop has a problem that it can hardly sample the most of the ground truth sufficiently. We ran a Monte-Carlo simulation to estimate the coverage of random crop on 12 epoch and characterized the probability of the coverage of the IoU of sampled region on the whole image. When we prolong the training epochs to 18, the coverage got a ascendance. Although the prolonging of the epochs can avoid the insufficient sampling somewhat, the distribution of coverage still keeps a fairly wide peak. 

\par We ran a same experiment with random seven crop. The following histogram is the result:

\begin{figure}[htb]

  \begin{minipage}[b]{0.45\linewidth}
    \centering
    \includegraphics[width = \linewidth]{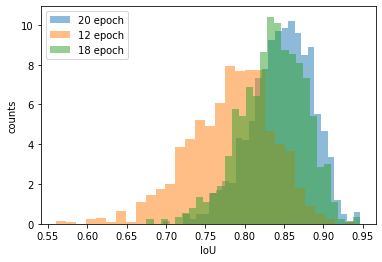}
  \end{minipage}\quad
  \begin{minipage}[b]{0.45\linewidth}
    \centering
    \includegraphics[width = \linewidth]{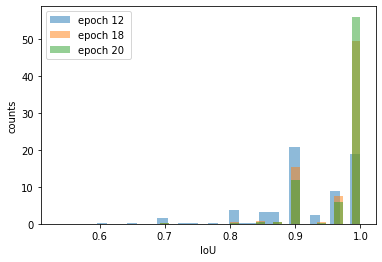}
  \end{minipage}\quad
\caption{Histogram of Crop Coverage: RandomCrop(left),\space RandomSevenCrop(right)}
\end{figure}

\par Then we ran the random seven crop with the average input size and the crop size as big as our GPU memory can afford.
\subsection{Cascade R-CNN}
The model can perform a considerable recall by the former operations of us. Ideally, precision is more important to raise the mmAP. The score of detection mmAP is calculated based on IoU, so we analyzed iou of all positive samples and plot a histogram as figure 5. 
\par Numbers and parameters of cascade head is determined by the experiment as table3.
\begin{table}[htb]
    \centering
    \begin{tabular}{ccc}
    \hline
         Head Number& IoU Threshold& mmAP \\
    \hline
         1 & 0.5 & 55.3 \\
         2 & [0.5, 0.6] &\textbf{56.9}\\
         3 &[0.5, 0.6, 0.7] & 56.7\\
    \hline
    \\
    \end{tabular}
    \caption{Parameter and Results of Cascade R-CNN}
    \label{tab:my_label}
\end{table}
\begin{figure}[htb]
    \centering
    \includegraphics[scale=0.6]{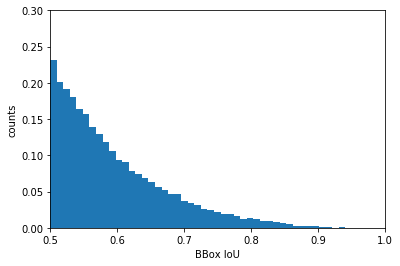}
    \caption{Positive Sample IoU of R-CNN}
    \label{fig:my_label}
\end{figure}

\subsection{Results}
Here is the list of adjustment used in the model with abbreviation:
\begin{itemize}
    \item ON: Optimal NMS - NMS with the parameters shown in 3.2
    \item OS: Optimal Sampler - The random sampler of RPN and R-CNN with optimal parameters
    \item OHEM - RPN sampler is the same with OS. R-CNN sampler is replaced by OHEM sampler\cite{ohem} instead.
    \item C: Cascade Head - The RCNN head cascaded with multi-head.
    \item ResNeXt - ResNeXt with 32 groups and base width of 4.
    \item BFP - BFP with 2 refine levels.

\end{itemize}
\begin{table}[htb]
    \centering
    \resizebox{150mm}{30mm}{
    \begin{tabular}{ccccccccccc}
    \hline
         ON& OS& C&Random Crop& Crop Size&Backbone&Neck& Input Size& Epoch &mmAP(\%) &mmAP(test)\\
    \hline
        &&&&&ResNet-50&FPN&(1333,800)&12& 50.6& \\
         \checkmark&&&&&ResNet-50&FPN&(1333,800)&12& \textbf{52.8}\\
         \checkmark&\checkmark&&&&ResNet-50&FPN&(1333,800)&12& \textbf{54.0}\\
         \checkmark&OHEM&&&&ResNet-50&FPN&(1333,800)&12&52.1 \\
         \checkmark&\checkmark&&Uniform&(800,800)&ResNet-50&FPN&(1333,800)&12&53.0\\
         \checkmark&\checkmark&&&&ResNet-50&FPN&(1800,1080)&12& \textbf{55.3}\\
         \checkmark&\checkmark&&Uniform&(1200,1200)&ResNet-50&FPN&No Rescale&12& 54.6\\
        \checkmark&\checkmark&&Uniform&(1200,1200)&ResNet-50&FPN&(3000,1800)&12& 55.3\\
        \checkmark&\checkmark&&Uniform&(1200,1200)&ResNet-50&FPN&(3000,1800)&18& \textbf{56.1}\\
        \checkmark&\checkmark&&Seven&(1200,1200)&ResNet-50&FPN&(3000,1800)&18& 55.1\\
        \checkmark&\checkmark&&Seven&(1200,1200)&ResNet-50&FPN&(3000,1800)&24& \textbf{56.6}\\
        \checkmark&\checkmark&\checkmark&Uniform&(1200,1200)&ResNet-50&FPN&(3000,1800)&18& \textbf{56.9}\\
        \checkmark&\checkmark&\checkmark&Seven&(1200,1200)&ResNet-50&FPN&(3000,1800)&24& \textbf{57.4}\\
        \checkmark&\checkmark&\checkmark&Seven&(1200,1200)&ResNet-50&BFP&(3000,1800)&24& \textbf{57.7}\\
        \checkmark&\checkmark&\checkmark&Seven&(1200,1200)&ResNeXt-101&BFP&(3000,1800)&24& \textbf{58.0}& \textbf{58.7}\\
    \hline
    \\
    \end{tabular}}
    \caption{Results and Conditions of All Experiment}
    \label{tab:my_label}
\end{table}
\clearpage
\bibliographystyle{unsrt}
\bibliography{template}

\end{document}